\documentclass{article}
\usepackage{spconf,amsmath,graphicx}
\usepackage{amsmath}
\usepackage{amssymb}
\usepackage{graphicx}
\usepackage[ruled,linesnumbered]{algorithm2e}
\usepackage{booktabs}
\usepackage{cuted}
\usepackage{balance}
\usepackage[hidelinks=true,bookmarks=FWLse]{hyperref}
\usepackage{lineno}
\usepackage{amsopn}
\usepackage{comment}
\usepackage{color}
\usepackage{cite}
\usepackage{algorithmic}
\usepackage{xcolor}
\usepackage{nomencl}
\usepackage{booktabs}
\usepackage{enumitem}
\usepackage{url,subfigure,tabulary}
\usepackage{bbding}

\newcolumntype{L}[1]{>{\raggedright\arraybackslash}p{#1}}
\newcolumntype{C}[1]{>{\centering\arraybackslash}p{#1}}
\newcolumntype{R}[1]{>{\raggedleft\arraybackslash}p{#1}}

\title{Multi-Scale Feature Fusion: Learning Better Semantic Segmentation for Road Pothole Detection}

\name{
		Jiahe Fan$^1$, Mohammud J. Bocus$^2$, Brett Hosking$^3$, Rigen Wu$^4$, 
		Yanan Liu$^2$, Sergey Vityazev$^5$, Rui Fan$^{6\star}$
}

\address{
$^1$Beijing Institute of Technology, Beijing 100811, P. R. China.\\
$^2$University of Bristol, Bristol, BS8 1TL, United Kingdom.\\
$^3$Arm, Manchester, M1 3HU, United Kingdom.\\
$^4$ATG Robotics, Hangzhou 310000, P. R. China.\\
$^5$Ryazan State Radio Engineering University, Ryazan 390005, Russia.\\
$^6$Tongji University, Shanghai 201804, P. R. China.\\  
Email: jhxfan@ieee.org, junaid.bocus@bris.ac.uk, brett.hosking@arm.com, wrg6370@outlook.com,\\
yl17692@bris.ac.uk, vityazev.s.v@ieee.org, rui.fan@ieee.org
\\
\thanks{$\star$ Corresponding Author}
}

\begin{document}
\maketitle

\begin{abstract}
This paper presents a novel pothole detection approach based on single-modal semantic segmentation. It first extracts visual features from input images using a convolutional neural network. A channel attention module then reweighs the channel features to enhance the consistency of different feature maps. Subsequently, we employ an atrous spatial pyramid pooling module (comprising of atrous convolutions in series, with progressive rates of dilation) to integrate the spatial context information. This helps better distinguish between potholes and undamaged road areas. Finally, the feature maps in the adjacent layers are fused using our proposed multi-scale feature fusion module. This further reduces the semantic gap between different feature channel layers. Extensive experiments were carried out on the Pothole-600 dataset to demonstrate the effectiveness of our proposed method. 
The quantitative comparisons suggest that our method achieves the state-of-the-art (SoTA) performance on both RGB images and transformed disparity images, outperforming three SoTA single-modal semantic segmentation networks. 
\end{abstract}


\section{Introduction}
Potholes are considerable structural failures on the road surface \cite{fan2019pothole}. They are caused by the contraction and expansion of the road surface as rainwater permeates the ground \cite{miller2003distress}. The affected road areas are further deteriorated due to tire vibration. This makes the road surface impracticable for driving \cite{mathavan2015review}. The vehicular traffic can cause subsurface materials to move, which further expands the potholes, creating a vicious circle \cite{fan2020we}. To avoid traffic accidents, it is crucial and necessary to detect road potholes in time \cite{fan2020rethinking}. With recent advances in machine learning, automated road pothole detection systems have become a reality \cite{wang2020applying, fan2019crack, koch2011pothole, fan2020sne}. Benefiting from the evolution of convolutional neural networks (CNNs), semantic segmentation has become an effective technique for road pothole detection \cite{fan2020rethinking}, and it has achieved compelling results.

Among the state-of-the-art (SoTA) semantic segmentation CNNs, fully convolutional network (FCN) \cite{long2015fully} replaces the fully connected layer used in traditional classification networks with a convolutional layer to achieve better segmentation results. Contextual information aggregation has proved to be an effective tool that can be used to improve segmentation accuracy. ParseNet \cite{liu2015parsenet} captures global context by concatenating global pooling features. PSPNet \cite{zhao2017pyramid} introduces a spatial pyramid pooling (SPP) module to collect contextual information in different scales. Atrous SPP (ASPP) \cite{chen2017deeplab,chen2017rethinking,chen2018encoder} applies different dilated convolutions to capture multi-scale contextual information without introducing extra parameters. 

To take advantage of global contextual visual information, some pioneering methods have been proposed to reweigh 2-D feature map channels. SE-Net \cite{hu2018squeeze} and EncNet \cite{zhang2018context} are designed to learn a globally-shared attention vector from global context. SE-Net \cite{hu2018squeeze} employs a squeeze-excitation operation to integrate the global contextual information into a feature weight vector and reweigh the feature maps. EcnNet \cite{zhang2018context} uses a context encoding module to obtain a globally-shared feature weight vector. This module adopts learning and residual encoding components to obtain a global context encoded feature vector, which is then used to predict the feature weight vector. Combining global context information to reweigh the feature map of each channel has proved to be effective in terms of improving semantic segmentation accuracy.

\begin{figure}[!t]
	\begin{center}
		\centering
		\includegraphics[width=0.475\textwidth]{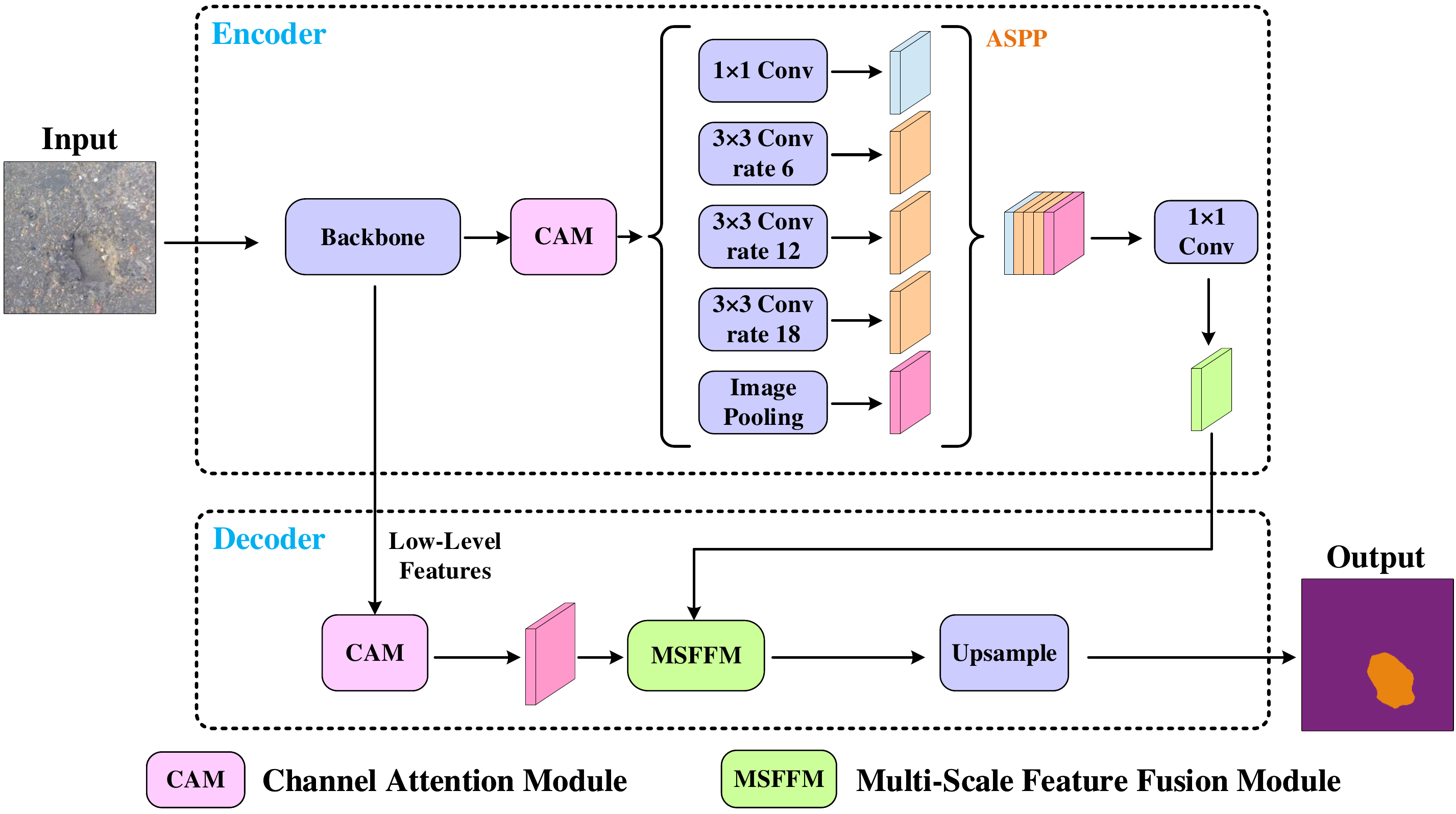}
		\centering
		\caption{The architecture of our proposed road pothole detection network.}
		\label{fig.block_diagram}
	\end{center}
\end{figure}

Some other methods use backbone CNNs \cite{zhao2017pyramid, zhang2018context, fu2019dual, chen2017rethinking} to extract feature maps at different scales. By performing a series of convolution and pooling operations, the top layer has rich semantic information \cite{chen2014semantic, eigen2015predicting, liu2015deep, liu2015learning}, while the lower-level feature maps contain fine-grained information \cite{lin2017refinenet}. This information asymmetry becomes a barrier to accurate semantic prediction. To address this issue, U-Net\cite{ronneberger2015u} adopts an encoder-decoder architecture to improve the semantic segmentation performance. It adds skip connections between the encoder and decoder, which can recover fine-grained details in the semantic prediction.  Feature pyramid network (FPN) \cite{lin2017feature} uses the structure of U-Net \cite{ronneberger2015u} with predictions from each level of the feature pyramid. However, the fusion operations cannot measure the semantic relevance between feature maps at different scales. The semantic information between feature maps at different scales may interfere with each other.

To address the above problems, in this paper, we propose a novel multi-scale feature fusion module (MSFFM) based on attention mechanism. Our main objective is to improve the semantic prediction by leveraging additional low-level information near the boundaries, where the pixel categories are difficult to infer. We utilize a matrix multiplication operation to measure the relevance between the two feature maps in the spatial dimension, which is the basic idea of weight vectors. By reweighing feature maps in lower layers, we reduce interference between feature maps in different layers. Moreover, we adopt a channel attention module (CAM) to reweigh feature maps in different channels to further improve the semantic segmentation results.

\section{Methodology}
Given a road image, potholes can have diverse shapes and scales. We can obtain feature maps at the top layer through a series of convolution and pooling operations. Although the feature maps have rich semantic information, their resolutions are not high enough to provide accurate semantic prediction. Unfortunately, directly combining low-level feature maps  can only bring very limited improvements. To  overcome this shortcoming, we design an effective feature fusion module in this paper. 

The schema of our proposed road pothole detection network is illustrated in Fig. \ref{fig.block_diagram}. Firstly, we employ a pre-trained dilated ResNet-101 as the backbone CNN to extract visual features. We also replace the down-sampling operations with dilated convolutions in the last two ResNet-101 \cite{he2016deep} blocks, thus the size of the final feature map is 1/8 of the input image. This module helps retain more details without introducing extra parameters. In addition, we adopt the ASPP module used in Deeplabv3 \cite{chen2017rethinking} to collect contextual information in the top feature map. Then, we adopt a CAM to reweigh the feature maps in different channels. It can highlight some features so as to produce better semantic predictions. Finally, we feed the feature maps at different levels into the MSFFM to improve the segmentation performance near the pothole contour.

\begin{figure}[!t]
	\begin{center}
		\centering
		\includegraphics[width=0.475\textwidth]{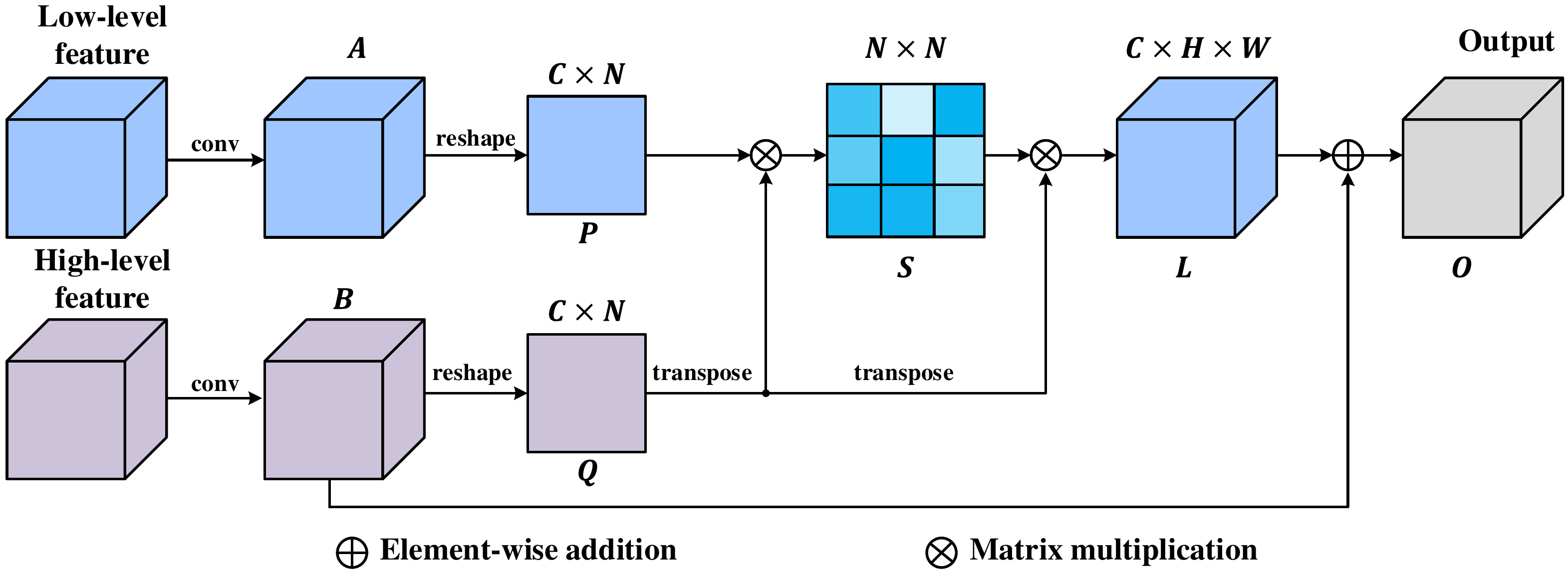}
		\centering
		\caption{Our proposed Multi-Scale Feature Fusion Module.}
		\label{fig.msffm}
	\end{center}
\end{figure}
\begin{figure}[!t]
	\begin{center}
		\centering
		\includegraphics[width=0.475\textwidth]{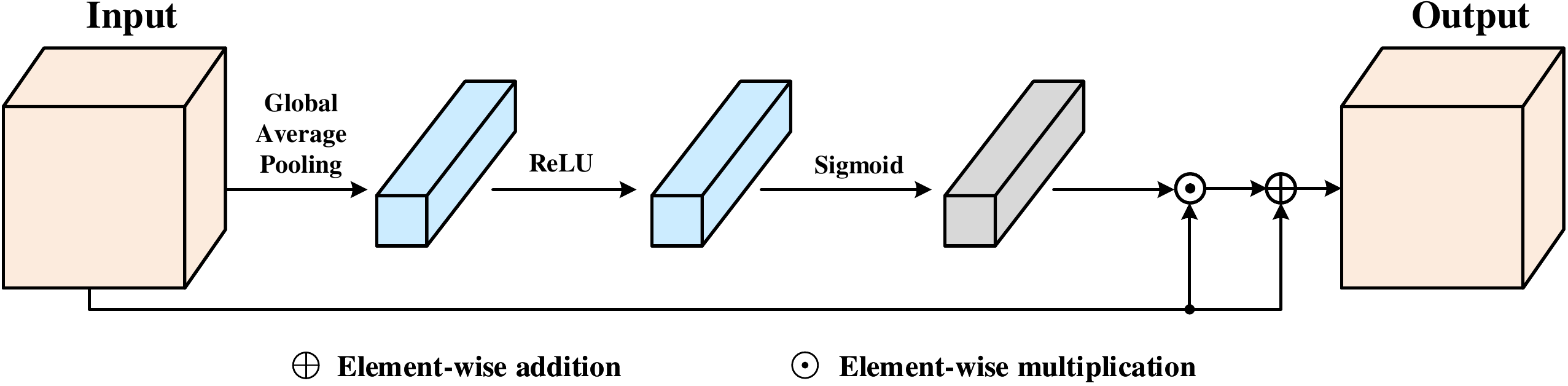}
		\centering
		\caption{Our employed Channel Attention Module. }
		\label{fig.CAM}
	\end{center}
\end{figure}

\subsection{Multi-scale feature fusion}\label{AA}
The top feature maps have rich semantic information but their resolution is low, especially near the pothole boundary. On the other hand, the lower feature maps have low-level semantic information but higher resolution. In order to address this problem, some works \cite{ronneberger2015u,chen2018encoder,badrinarayanan2017segnet} directly combine the feature maps in different layers. Nevertheless, their achieved improvements are very limited because of the semantic gap between feature maps with different scales.

The attention modules have been widely applied in many works \cite{lin2017structured, vaswani2017attention, shen2018disan}. Inspired by some successfully applied spatial attention mechanisms, we introduce a MSFFM, which is based on spatial attention to efficiently fuse the feature maps at different scales. Semantic gap is one of the key challenges in feature fusion. To solve this issue, the MSFFM calculates the correlation between pixels in different feature maps via matrix multiplication, and the correlation is then utilized as the weight vectors for the higher-level feature map:

\begin{equation}
	{{\rm{s}}_{ji}} = \frac{{\exp ({P_i} \cdot {Q_j})}}{{\sum\nolimits_{i = 1}^N {\exp ({P_i} \cdot {Q_j})} }},
	\label{eq.msffm}
\end{equation}
where ${s_{ji}}$ measures the relevance between the $i$-$th$ position in lower feature map and the $j$-$th$ position in higher feature map. $N$ represents the number of pixels. $P$ and $Q$ represent the lower and higher feature maps generated by the convolutional layer, respectively, where $\{P, Q\}\in{{\mathbb{R}}^{C\times N}}$. The higher the similarity between feature representations of pixels at the two positions, the greater is the relevance between them. As shown in Fig. \ref{fig.msffm}, we first feed the feature maps into a convolution layer to compress the channels for fewer calculations while generating feature maps $A$ and $B$, $\{A,B\}\in{{\mathbb{R}}^{C\times H\times W}}$. $H$ and $W$ represent the height and width of the feature map. Then we reshape the low-level feature map $A$ and the high-level feature map $B$ to $P$ and $Q$, respectively, where $N=H\times W$ represents the number of pixels. Afterwards, we transpose $Q$ for matrix multiplication and apply a softmax layer to calculate the spatial attention map  $S \in {{\mathbb{R}}^{N \times N}}$.

Then we perform matrix multiplication between $Q$ and the spatial attention map $S$ to generate the feature map $L\in{{\mathbb{R}}^{C\times H\times W}}$. Finally, we utilize an element-wise sum operation between $B$ and $L$ to obtain the final output $O\in{{\mathbb{R}}^{C\times H\times W}}$ as follows:

\begin{equation}
	{O_j} = \alpha \sum\limits_{i = 1}^N {({s_{ji}}{{\rm{q}}_{{i}}}) + } {B_j},
\end{equation}
where $\alpha$ is initialized as 0 and it gradually learns to assign more weight, ${{\rm{q}}_i}$ represents the $i$-$th$ position in the lower feature map, and ${B_j}$ represents the $j$-$th$ channel of the top feature map. It can be inferred from (2) that each position of the final feature $O$ is a weighted sum of the features across all positions of the top features. As the final feature is generated by the top features, the high-level semantic information is well preserved in the final outputs. 

In summary, we utilize matrix multiplication to measure the relevance of pixels in feature maps from different layers, which integrates the detailed information from the lower feature map into the final outputs, thus improving the semantic segmentation performance for the pothole boundary. We apply this module between the last two layers.

\subsection{Channel-wise feature reweighing}\label{BB}
It is well-known that high-level features have rich semantic information and each channel map can be regarded as a class-specific response. Each response can affect the final semantic prediction to a different extent. Therefore, we utilize CAMs, as illustrated in Fig. 3, to enhance the consistency of the feature maps in each layer, by changing the features' weights in each channel. The CAM is designed to reweigh each channel according to the overall pixels of each feature map. We first employ a global average pooling layer to squeeze spatial information. Subsequently, we use the Rectified Linear Unit (ReLU) and sigmoid function to generate the weight vectors, which are finally combined with the input feature maps by element-wise multiplication operations to generate an output feature map. The overall information is integrated into the weight vectors, making the feature maps more reliable and the pothole detection results closer to the ground truth. In our experiments, we employ the CAM in the 4th and 5th layers.

\begin{figure}[!t]
	\begin{center}
		\centering
		\includegraphics[width=0.475\textwidth]{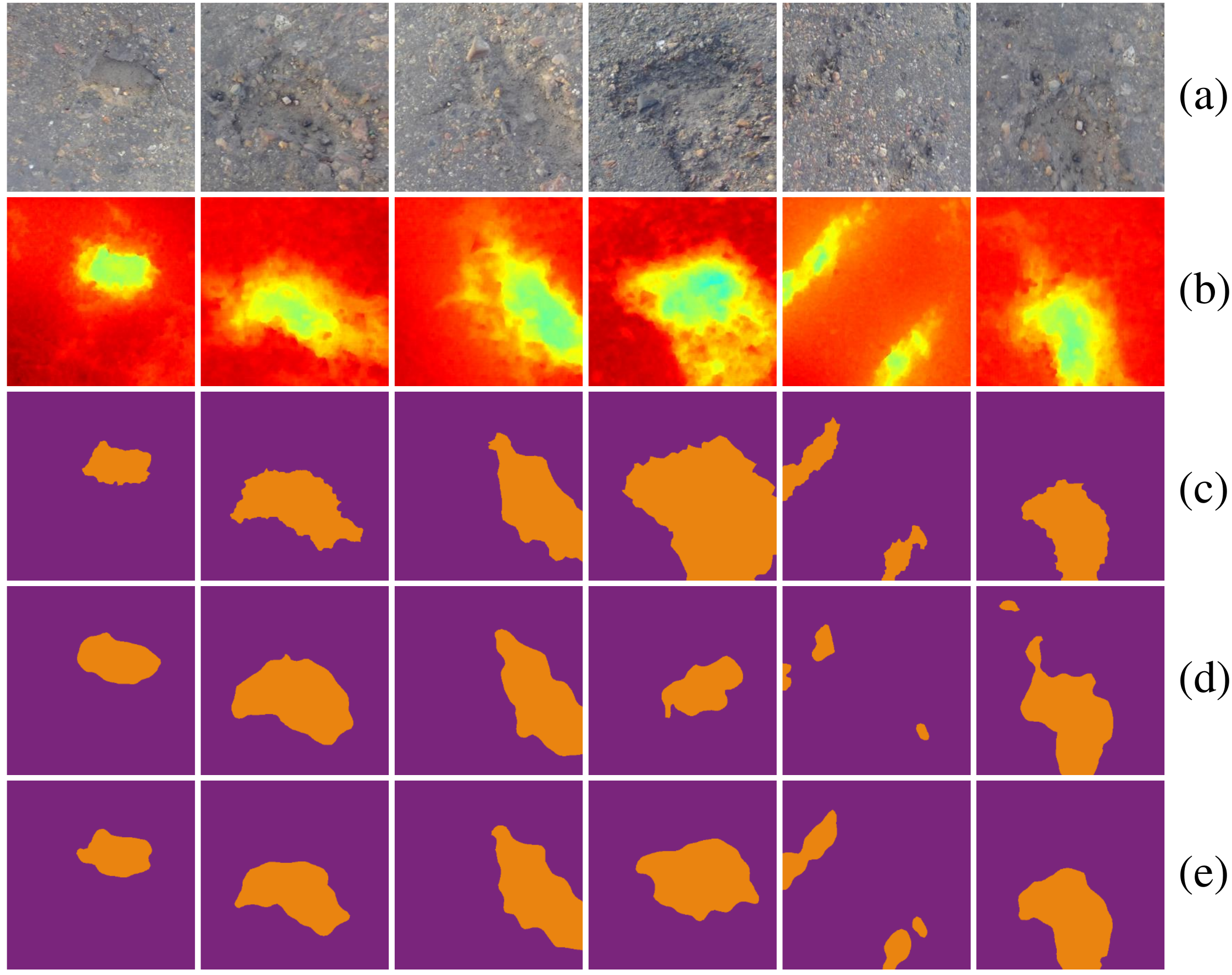}
		\centering
		\caption{Examples of pothole detection results: (a) RGB images; (b) transformed disparity images; (c) pothole  ground truth; (d) semantic RGB image segmentation results; (e) semantic transformed disparity image segmentation results.}
		\label{fig.results}
	\end{center}
\end{figure}

\section{Experiment Results}
In this paper, we carry out comprehensive experiments on the Pothole-600 dataset \cite{fan2020we} to evaluate the performance of our proposed road pothole detection both qualitatively and quantitatively. This dataset provides two modalities of vision sensor data: 1) RGB images, and 2) transformed disparity images \cite{fan2019real}. The transformed disparity images were obtained by performing disparity transformation \cite{fan2019road, wang2021dynamic} on dense disparity images estimated by PT-SRP \cite{fan2018road}. We conduct experiments to select the best architecture. All the experiments use the same training setups.

\textbf{Ablation Study}: To validate the effectiveness of our proposed MSFFM and CAM, we first carry out the ablation study on different network architectures, as shown in Table 1 and Table 2. The baseline network uses Deeplabv3\cite{chen2017rethinking}, which concatenates the feature maps from ASPP module and the lower layer.
 
Moreover, we implement the two modules into the baseline network and verify their effectiveness, respectively. According to the results shown in Table 1 and Table 2, implementing two modules can achieve better performance than the baseline network on both RGB images and transformed disparity images. The mIoU improvements on RGB images with the use of CAM and MSFFM are 1.85$\%$ and 4.11$\%$, respectively, while the mIoU improvements on the transformed disparity images are 1.36$\%$ and 0.12$\%$, respectively. The network with MSFFM and CAM embedded yields an mFsc of 76.16$\%$ on RGB images and an mFsc of 84.22$\%$ on transformed disparity images. Based on these experimental results, we believe that the CAM and MSFFM adopted in our network can improve the segmentation accuracy significantly.

\begin{table}[!t]
	\begin{center}
		\footnotesize
		\caption{Ablation study on RGB images. }
		\begin{tabular}{L{3.9cm}|C{1.4cm}C{1.4cm}}
			\toprule
			Methods & {mIoU ($\%$)} & {mFsc ($\%$)} \\
			\hline
			Baseline &55.32 &71.23 \\
			Baseline + CAM &57.17 &72.75 \\
			Baseline + MSFFM &59.43 &74.55 \\
			Baseline + CAM + MSFFM (ours) &61.51 &76.16 \\
			\bottomrule
		\end{tabular}
		\label{tab.detection}
	\end{center}
	\vspace{-2em}
\end{table}

\begin{table}[!t]
	\begin{center}
		\footnotesize
		\caption{Ablation study on transformed disparity images. }
		\begin{tabular}{L{3.9cm}|C{1.4cm}C{1.4cm}}
			\toprule
			Methods & {mIoU ($\%$)} & {mFsc ($\%$)} \\
			\hline
			Baseline &70.90 &82.97 \\
			Baseline + CAM &72.26 &83.89 \\
			Baseline + MSFFM &71.02 &83.06 \\
			Baseline + CAM + MSFFM (ours) &72.75 &84.22 \\
			\bottomrule
		\end{tabular}
		\label{tab.detection}
	\end{center}
	\vspace{-2em}
\end{table}

\begin{table}[!t]
	\begin{center}
		\footnotesize
		\caption{Performance of other SoTA networks on RGB images. }
		\begin{tabular}{L{2.4cm}|C{2.4cm}C{2.4cm}}
			\toprule
			Methods & {mIoU ($\%$)} & {mFsc ($\%$)} \\
			\hline
			PSPNet \cite{zhao2017pyramid} &58.61 &73.90 \\
			DANet \cite{fu2019dual} &59.42 &74.54 \\
			Deeplabv3 \cite{chen2018encoder} &58.60 &73.90 \\
			\bottomrule
		\end{tabular}
		\label{tab.rgb}
	\end{center}
	\vspace{-2em}
\end{table}

\begin{table}[!t]
	\begin{center}
		\footnotesize
		\caption{Performance of other SoTA networks on transformed disparity images.}
		\begin{tabular}{L{2.4cm}|C{2.4cm}C{2.4cm}}
			\toprule
			Methods & {mIoU ($\%$)} & {mFsc ($\%$)} \\
			\hline
			PSPNet \cite{zhao2017pyramid} &69.85 & 82.25 \\
			DANet \cite{fu2019dual} &70.52 & 82.71\\
			Deeplabv3 \cite{chen2018encoder} &70.36 &82.60 \\
			\bottomrule
		\end{tabular}
		\label{tab.tr}
	\end{center}
	\vspace{-2em}
\end{table}

\textbf{Performance Comparison}: We also compare our method with three SoTA semantic segmentation CNNs: 1) Deeplabv3 \cite{chen2018encoder}, 2) PSPNet \cite{zhao2017pyramid}, 3) DANet \cite{fu2019dual} on both RGB images and transformed disparity images, as shown in Table \ref{tab.rgb} and Table \ref{tab.tr}. PSPNet \cite{zhao2017pyramid} and Deeplabv3 \cite{chen2018encoder} collect contextual information in different scales, and therefore, they achieve similar results on RGB images and transformed disparity images. DANet \cite{fu2019dual} collects contextual information based on attention mechanism and it shows better performance on both RGB images and transformed disparity images. This further demonstrates the superiority of attention mechanism on semantic segmentation for road pothole detection, which can also be observed from the comparison between our method and other SoTA networks. 

Additionally, when using RGB images, the mIoUs of our method are 2.91$\%$, 2.9$\%$, and 2.09$\%$ higher than those achieved by Deeplabv3 \cite{chen2018encoder}, PSPNet \cite{zhao2017pyramid}, and DANet \cite{fu2019dual}, respectively. Moreover, our method also outperforms the above-mentioned SoTA semantic segmentation networks on transformed disparity images, where the improvements on mIoU with respect to Deeplabv3 \cite{chen2018encoder}, PSPNet \cite{zhao2017pyramid}, and DANet \cite{fu2019dual} are 2.39$\%$, 2.9$\%$, and 2.23$\%$, respectively. Specifically, our method achieves the best performance, even when it  only utilizes a MSFFM.

We also provide some qualitative results of our proposed road pothole detection method in Fig. \ref{fig.results}, where it can be observed that the CNN achieves accurate results on the transformed disparity images. The results obtained from our comprehensive experimental evaluations have demonstrated the effectiveness and superiority of our method compared to other SoTA techniques. Owing to the proposed CAM and MSFFM, our method achieves better performance for potholes detection on both RGB and transformed disparity images.

\section{Conclusion}
This paper introduced a method to detect road potholes based on semantic segmentation, which employs a novel multi-scale feature fusion module based on spatial attention to reduce the semantic gap between the feature maps in different layers. This helps maintain the semantic information in the higher-level feature maps and combine the detailed information near the pothole boundary. The top feature maps can be reweighed using the vectors generated by the relevance of each pixel in the different layers, which combine the global information of the feature maps. Moreover, a channel attention module is introduced to strengthen the channels which are more relevant to the  semantic segmentation ground truth. Extensive experiments were conducted on both RGB images and transformed disparity images, where our proposed network outperforms all other SoTA semantic segmentation networks.

\bibliographystyle{IEEEbib}

\end{document}